\def\year{2021}\relax
\documentclass[letterpaper]{article} 
\usepackage{aaai21}  
\usepackage{times}  
\usepackage{helvet} 
\usepackage{courier}  
\usepackage[hyphens]{url}  
\usepackage{graphicx} 
\usepackage{natbib}  
\usepackage{caption} 
\usepackage[switch]{lineno}  %

\usepackage{array}
\usepackage{amsmath}
\usepackage{float}
\usepackage{booktabs}
\usepackage{multirow}
\usepackage{stfloats}
\usepackage{color}
\usepackage{amsfonts}
\newtheorem{theorem}{Assumption}

\usepackage{bm}
\usepackage{enumerate}
\usepackage{threeparttable}
\usepackage{algorithm}
\usepackage{algpseudocode}

\title{Distribution Adaptive INT8 Quantization for Training CNNs}
\author {
Kang Zhao, Sida Huang, Pan Pan, Yinghan Li, Yingya Zhang, Zhenyu Gu, Yinghui Xu

}
\affiliations {
Machine Intelligence Technology Lab, Alibaba Group \\
\{zhaokang.zk, sida.hsd, panpan.pp, lyh238099, yingya.zyy, zhenyu.gu, renji.xyh\}@alibaba-inc.com
}
\begin{document}

\maketitle

\begin{abstract}
Researches have demonstrated that low bit-width (e.g., INT8) quantization can be employed to accelerate the inference process. It makes the gradient quantization very promising since the backward propagation requires approximately twice more computation than forward one. Due to the variability and uncertainty of gradient distribution, a lot of methods have been proposed to attain training stability. However, most of them ignore the channel-wise gradient distributions and the impact of gradients with different magnitudes, resulting in the degradation of final accuracy. In this paper, we propose a novel INT8 quantization training framework for convolutional neural network to address the above issues. Specifically, we adopt \emph{Gradient Vectorized Quantization} to quantize the gradient, based on the observation that layer-wise gradients contain multiple distributions along the channel dimension. Then, \emph{Magnitude-aware Clipping Strategy} is introduced by taking the magnitudes of gradients into consideration when minimizing the quantization error, and we present a theoretical derivation to solve the quantization parameters of different distributions. Experimental results on broad range of computer vision tasks, such as image classification, object detection and video classification, demonstrate that the proposed Distribution Adaptive INT8 Quantization training method has achieved almost lossless training accuracy for different backbones, including ResNet, MobileNetV2, InceptionV3, VGG and AlexNet, which is superior to the state-of-the-art techniques. Moreover, we further implement the INT8 kernel that can accelerate the training iteration more than 200\% under the latest Turing architecture, i.e., our method excels on both training accuracy and speed.
\end{abstract}

\section{Introduction}

\begin{figure}[h] 
	\centering
	\includegraphics[width=0.48\textwidth]{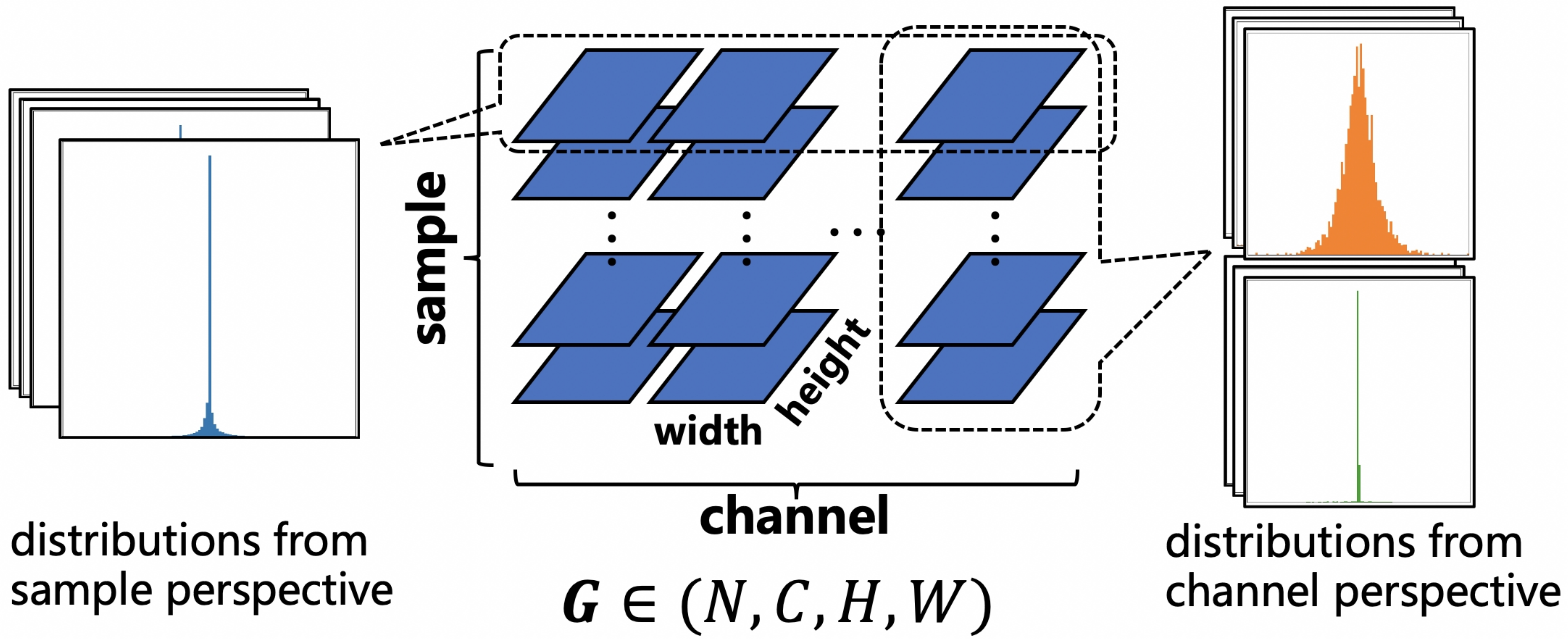} 
	\caption{The observation of layer-wise gradient distribution from sample and channel perspectives, with respect to conv6, 21, 34, 46 of ResNet-50 on ImageNet. It is obvious there exist two categories of gradient distribution observed along channel dimension: a bell-shaped distribution, and a sharp with long-tailed shape distribution. But the distributions observed from sample perspective look similar.} 
	\label{fig.channelperspective} 
\end{figure}

Many literatures \cite{jacob2018quantization,wang2018two,hubara2016binarized,choi2019accurate,jain2019trained} have demonstrated that low bit-width quantization can be utilized to speed up the inference stage (weight or activation) with minor hurt of accuracy. It makes quantization of gradient quite promising since the backward propagation consumes twice more computing resources than forward propagation \cite{banner2018scalable}. Among the family of low bit-width formats (e.g., FP16, INT16, FP8, INT8, INT4, etc), INT8 is widely adopted in community for the trade-off between efficiency and accuracy. On the one hand, INT8 operation is twice faster than FP16/INT16, and supported by more architectures compared with FP8. On the other hand, INT8 has a larger quantization range (-128$\sim$127) than INT4 (-8$\sim$7) or other lower bit-widths (less than 4-bits).

Although quantizing gradient with INT8 to accelerate training is appealing, there still remains a big challenge, because the quantization error of gradient easily misleads the direction of convergence as analyzed in \cite{zhu2020towards}. Many methods have been proposed to keep the training process reliable, most of them either use floating point numbers to simulate INT8 calculation, which has no actual acceleration effect, or resort to inappropriate quantization schemes that result in large deterioration of final accuracy. \cite{zhu2020towards} adopts a unified solution to make INT8 training stable for a variety of networks, but it still suffers from some loss of accuracy (up to 1\% on ImageNet), which is not acceptable in practice. Combining INT16 with INT8 \cite{zhang2020fixed} is another option to compensate the quantization loss of gradient but outside the scope of our work.

In this paper, we propose a Distribution Adaptive INT8 Quantization training framework from the aspects of training accuracy and speedup. The main contributions of our work are outlined as follows:
\begin{itemize}
	\item Based on an important observation ignored by previous works that the layer-wise gradients contains multiple distributions, we employ \emph{Gradient Vectorized Quantization} to well capture the gradient distribution.
	\item We propose \emph{Magnitude-aware Clipping Strategy} which considers the impact of gradients with different magnitudes when minimizing quantization error, and present a theoretical derivation to solve the optimal quantization parameters of different distributions.
	\item Under the latest Turing architecture, we implement TensorCore\footnote{\url{https://www.nvidia.com/en-us/data-center/tensorcore/}}-based INT8 kernel, which achieves 18\% faster than FP16 and speeds up FP32 more than 200\%.
	\item To the best of our knowledge, we are the first INT8 training framework to achieve almost lossless performance compared with full precision (FP32) on a variety of tasks and backbones.
\end{itemize}

\section{Related Work}
Generally speaking, quantization methods can be roughly divided into two major categories: inference quantization and training quantization, according to distinct quantization targets.

\textbf{Inference Quantization.} Inference quantization aims to quantize weight and/or activation to accelerate the forward pass. BNNs \cite{hubara2016binarized} trains the neural networks with binary weight and activation successfully. TWNs \cite{li2016ternary} constrains weight to +1, 0 and -1 and optimizes a threshold-based ternary function. These low bit-widths (less than 4-bits) quantization techniques have significant degradation in performance. To improve the accuracy, more advanced quantization schemes \cite{Cai_2017,zhang2018lq,jung2019learning} have been proposed. In \cite{choi2019accurate}, the combination of range ReLU and statistics-aware weight binning make the 2-bit quantization get comparable performance with full precision networks. By training quantization thresholds, TQT \cite{jain2019trained} also achieves near-floating-point accuracy.

\textbf{Training Quantization.} Compared with inference quantization, training quantization needs to further quantize the gradient to accelerate the whole training process. \cite{banner2018scalable} replaces batch-norm (BN) layer with RangeBN and adopt ``Gradients Bifurcation'' to quantize both gradient and BN layers to 8-bit, but it de-quantizes gradient to floating point number before calculation, which can not be speed up by INT8 operator. A similar case is \cite{wang2018training}, which needs new hardware platforms to boost the speed. Besides quantizing gradient to INT8, WAGE \cite{wu2018training} and WAGEUBN \cite{yang2020training} further quantize other training modules (e.g., Error and Update), leading to large loss of accuracy. \cite{zhu2020towards} utilizes direction sensitive gradient clipping and counteractive learning rate scaling to guide the 8-bit gradient learning and updating, which help it succeed in many networks and tasks.

Considering the INT8 quantization of weight and activation have been well studied in \cite{zhu2020towards, zhang2020fixed}, we will take the same way to quantize weight and activation as they did, and concentrate on the gradient quantization in the rest of the paper.

\section{Our Method}
This section describes the formulation of our INT8 quantization framework. 
To facilitate our discussion, the quantization method is given below.

Following \cite{zhu2020towards, zhang2020fixed}, we employ the symmetric uniform quantization. For a given data $x$, and a quantization parameter $s$, the fixed point 8-bit integer quantization is defined as: 
\begin{eqnarray}\label{eqn.suq}
q(x) = round(127 \cdot \frac{clamp(x, s)}{s}),
\end{eqnarray}
where $clamp(.)$ is a clipping function:
\begin{eqnarray}
clamp(x, s) = \left\{
\begin{array}{lcl}
x, & \left | x \right | \le s \\
sign(x) \cdot s, &  \left | x \right | > s \\
\end{array}.
\right.
\end{eqnarray}
It can guarantee $x$ in the range of [-s, s]. Correspondingly, the de-quantization of $x$ can be easily expressed as: $\hat{x} = q(x) \times \frac{s}{127}$. Similar to prevailing methods, we utilize stochastic rounding \cite{gupta2015deep} for gradient quantization and nearest rounding for weight and activation quantization. 

\subsection{Motivation}
Clearly, $s$ is critical to the quantization error of gradient, since it is the only variable in Eq.(\ref{eqn.suq}). The existing approaches either simply take $s$ as $max(|x|)$ or minimize the mean-square error (MSE) or cosine distance between $x$ and $\hat{x}$ to find a better value of $s$. They all encounter the following two issues: \\
i) \textbf{The multiple distributions of the layer-wise gradients are not taken into account}. In order to minimize the quantization error, the previous ways \cite{banner2018scalable,he2019simultaneously,zhu2020towards,zhang2020fixed} try to find a global value of $s$ by assuming that the gradient satisfies a certain distribution, or adopt periodic gradient descent trick for seeking an optimal $s$ without distribution assumption. These methods only use one global quantization parameter ($s$) for each layer's gradients (we call it global quantization for short). Based on our observations, we found that there exist more than one distribution of gradients inside one layer. And different distributions usually have different optimal quantization parameters. For example, taking $s$ as $max(|x|)$ is good enough for Gaussian Distribution, yet Laplacian Distribution will prefer a $s$ smaller than $max(|x|)$ in consideration of the densely distributed gradients that are far away from the maximum. It indicates multiple quantization parameters can capture the gradient distribution better than only one quantization parameter.\\
ii) \textbf{The contribution of gradients with different magnitudes is not considered}. \cite{lin2017deep,strom2015scalable} show that it is more important to transmit large gradients than small ones during the communication process, because large gradients contain more information. Similarly, the quantization error of large gradients should be more important than small ones in view of training accuracy. As stated in \cite{banner2018scalable,zhu2020towards}, small values will take the majority of gradients with training evolving, which means if we ignore the magnitudes of gradients, the quantization error and quantization parameter will be dominated by the small gradients. In that case, the quantization error of large gradients will be larger, leading to the deterioration of final accuracy.

To solve these two issues, we propose \emph{Gradient Vectorized Quantization} and \emph{Magnitude-aware Clipping Strategy}, respectively.

\subsection{Gradient Vectorized Quantization}
Without loss of generality, we assume the input of a convolution layer is $\mathbf{X} \in (N,C_{in},H_{in},W_{in})$\footnote{$(N,C_{in},H_{in},W_{in})$ is short for $\mathrm{R}^{N \times C_{in} \times H_{in} \times W_{in}}$}. $\mathbf{W} \in (C_{out},C_{in},k_1,k_2)$ is the weight parameter and the output is $\mathbf{Y} \in (N,C_{out},H_{out},W_{out})$. During backward propagation, the gradients of $\mathbf{Y}$ is denoted as $\mathbf{G}$. We observe $\mathbf{G}$ of each layer from both sample and channel perspectives during the training process, as shown in Figure \ref{fig.channelperspective}. It is found that the channel dimension obviously split the distribution of $\mathbf{G}$ into two categories: a bell-shaped distribution, and a sharp with long-tailed shape distribution (we call it Inverted-T distribution). Similar phenomena is not found from the sample dimension, which may arise from the specific attributes contained in channel dimension. Considering we will use different quantization parameters for these two distributions (details shown in the next section), we apply \emph{Gradient Vectorized Quantization} along the channel dimension.

In the gradient calculation of convolution layer, $\mathbf{G}$ is coming from the previous layer, then $\mathbf{G}_X$ and $\mathbf{G}_W$ can be calculated as following:
\begin{align}
\mathbf{G}_X &= \mathbf{G}\odot \mathbf{W} \label{eqn.gx}, \\
\mathbf{G}_W &= \mathbf{X}^{\prime} \otimes \mathbf{G}^{\prime}, \label{eqn.gw}
\end{align}
where $\odot$ and $\otimes$ are two specific convolution operations\footnote{$\odot$ is transpose convolution, and $\otimes$ is dilation convolution in practice}. In the implementation of Eq.(\ref{eqn.gw}), we usually do a dimension transformation on $\mathbf{X}$ and $\mathbf{G}$ to make the shape match: $\mathbf{X} \rightarrow \mathbf{X}^{\prime} \in (C_{in},N,H_{in},W_{in}), \mathbf{G} \rightarrow \mathbf{G}^{\prime} \in (C_{out},N,H_{out},W_{out})$.

Now, we consider the vectorized quantization of Eq.(\ref{eqn.gw}). $q(\mathbf{X}^{\prime})$ can be easily obtained by making a similar dimension transformation of $q(\mathbf{X})$ (it has been calculated in forward pass), and its quantization parameter is denoted as $s_x$. Let $\mathbf{G}^{\prime} = [\mathbf{G}_1^{\prime},...,\mathbf{G}_i^{\prime},...,\mathbf{G}_{C_{out}}^{\prime}]^\top$, where $\mathbf{G}_i^{\prime} \in (N,H_{out},W_{out})$, we quantize $\{\mathbf{G}_i^{\prime}\}_{i=1}^{C_{out}}$ with different quantization parameter $\{s_i\}_{i=1}^{C_{out}}$ (how to get $s_i$ will be described in the following section): 
\begin{eqnarray}
q(\mathbf{G}^{\prime}) = [q(\mathbf{G}_1^{\prime}),...,q(\mathbf{G}_i^{\prime}),...,q(\mathbf{G}_{C_{out}}^{\prime})]^\top.
\end{eqnarray}
Then we have:
\begin{align}
q(\mathbf{G}_W) &= q(\mathbf{X}^{\prime}) \otimes q(\mathbf{G}^{\prime}) \nonumber \\
&=  [q(\mathbf{X}^{\prime}) \otimes q(\mathbf{G}_1^{\prime}),...,q(\mathbf{X}^{\prime}) \otimes q(\mathbf{G}_i^{\prime}),...]^\top \nonumber \\
&= [q(\mathbf{G}_{W_1}),...,q(\mathbf{G}_{W_i}),...,q(\mathbf{G}_{W_{C_{out}}})] ^\top \nonumber,
\end{align}
where $q(\mathbf{G}_{W_i}) = q(\mathbf{X}^{\prime}) \otimes q(\mathbf{G}_i^{\prime})$ with the shape size $(C_{in},k_1,k_2)$. The de-quantization of $\mathbf{G}_W$ is also very simple:
\begin{align}
\hat{\mathbf{G}}_W &= [\hat{\mathbf{G}}_{W_1},...,\hat{\mathbf{G}}_{W_i},...,\hat{\mathbf{G}}_{W_{C_{out}}}]^\top, \\
\hat{\mathbf{G}}_{W_i} &= q(\mathbf{G}_{W_i}) \cdot \frac{s_x}{127} \cdot \frac{s_i}{127},
\end{align}
which can be implemented by an elementary product operation in PyTorch framework.

We use global quantization (as we did in the forward pass) to quantize Eq.(\ref{eqn.gx}), since there is no obvious difference in the distribution of $\mathbf{G}$ from sample dimension, which is also consistent with \cite{banner2018scalable} that $\mathbf{G}_W$ need higher quantization accuracy than $\mathbf{G}_X$.

Noted that although vectorized quantization has $C$ (numbers of channel) quantization parameters, it has the same computational complexity as global quantization. Take seeking the quantization parameter as an example, global quantization requires searching $N \times C \times H \times W $ float spaces to find the optimal $s$, while vectorized quantization only needs to search $N \times H \times W$ float spaces for each quantized parameter, which is basically equal on the whole.

\subsection{Magnitude-aware Clipping Strategy}
Because large gradients contribute more on network accuracy than smaller ones, we introduce the magnitudes of gradients ($f(g)$) into quantization error as follows:
\begin{align}
E &= \int_{g_{min}}^{g_{max}}\left | g-\hat{g} \right |f(g)p(g)dg \label{eqn.e},
\end{align}
where $g$ is the continuous random variable in the range of $[g_{min},g_{max}]$ for $\{\mathbf{G}_i^{\prime}\}_{i=1}^{C_{out}}$, $\hat{g}$ stands for its corresponding de-quantization, and $p(g)$ denotes the gradient distribution.

Mathematically, the positive correlation between magnitude contribution and gradient amplitude ($\left | g \right |$) can be expressed as $f(g) = e^{\alpha \left | g \right |} $. Note that nonnegative $\alpha$ reflects the importance of large amplitude to quantization error. More concretely, for two gradients $g_{1}$, $g_{2}$ with $\left | g_1 \right | < \left | g_2 \right |$, $frac\left( E, g_2 \right) $ \footnote{$frac\left( E, x \right)=\frac{1}{E} \int_{x}^{x+\epsilon}\left | g-\hat{g} \right |f(g)p(g)dg$, where $\epsilon$ is an infinitesimal value} becomes much greater than $frac\left( E, g_1 \right) $ with the increase of $\alpha$.

Considering the complexity of Eq.(\ref{eqn.e}) and difficult-to-measure of $\alpha$, we make the following two assumptions to grasp the key features in choosing quantization parameter: 
\begin{theorem}
	$\forall g\in \mathbb{R}, p(g)=p(-g)$; \label{asump.sym}
\end{theorem}

\begin{theorem}
	$\forall g\in (0,s), \left | g - \hat{g} \right |\approx \frac{1}{2}\cdot \frac{s}{127}$. \label{asump.quan}
\end{theorem}
The first assumption is based on our observation that the gradient distribution is nearly symmetric around the origin. And the second one utilizes the rectangle rule, which is widely adopted in numerical integration \cite{mccallum2012calculus}. According to the above assumptions, Eq.(\ref{eqn.e}) is simplified to:
\begin{align}
E &= \underset{I1}{\underbrace{\int_{0}^{s}\frac{s}{127}\cdot f(g)p(g)dg}}+\underset{I2}{\underbrace{2\int_{s}^{g_{max}}(g-s)\cdot f(g)p(g)dg}}. \label{eqn.e2}
\end{align}

Eq.(\ref{eqn.e2}) quantitatively describes the contradictory effect of $s$ on $E$: as $s$ goes down, $I1$ decreases yet $I2$ increases, and vice versa. To seek a balance between $I1$ and $I2$, the optimal quantization parameter $\bar{s}$ should satisfy the following condition: 
\begin{align}
\left. \frac{\partial E}{\partial s} \right |_{s=\bar{s}}=0 \label{eqn.deri1}.
\end{align}
Next, we discuss how to choose $\bar{s}$ under different gradient distributions.

\subsubsection{Gaussian distribution.} We might as well assume the bell-shaped gradients can be described by Gaussian distribution, which takes the max absolute value ($\left | g \right |_{max}$) as optimal quantization parameter \cite{banner2018scalable}:
\begin{align}
\bar{s} &= \left | g \right |_{max} \label{eqn.gaus}.
\end{align}

In general, the quantization parameter can be denoted by: $s = \beta \left | g \right |_{max}$, with $\beta \in \left( 0,1\right] $. There exists the following inequality constraint between $\alpha$ and $\beta$:
\begin{align}
\frac{\partial }{\partial \alpha}\left ( \frac{\partial E}{\partial s} \right )\cdot 
\frac{\partial }{\partial \beta}\left ( \frac{\partial E}{\partial s} \right )< 0. \label{eqn.inequal}
\end{align}
Eq.(\ref{eqn.inequal}) suggests that $\beta$ increases as $\alpha$ grows. The increase of $\beta$ makes $s$ closer to $\left | g \right |_{max}$, so we reuse Eq.(\ref{eqn.gaus}) to get quantization parameter of Gaussian distribution after considering the magnitude contribution ($\alpha$).

\begin{figure*} [t]
	\centering
	\includegraphics[width=\textwidth]{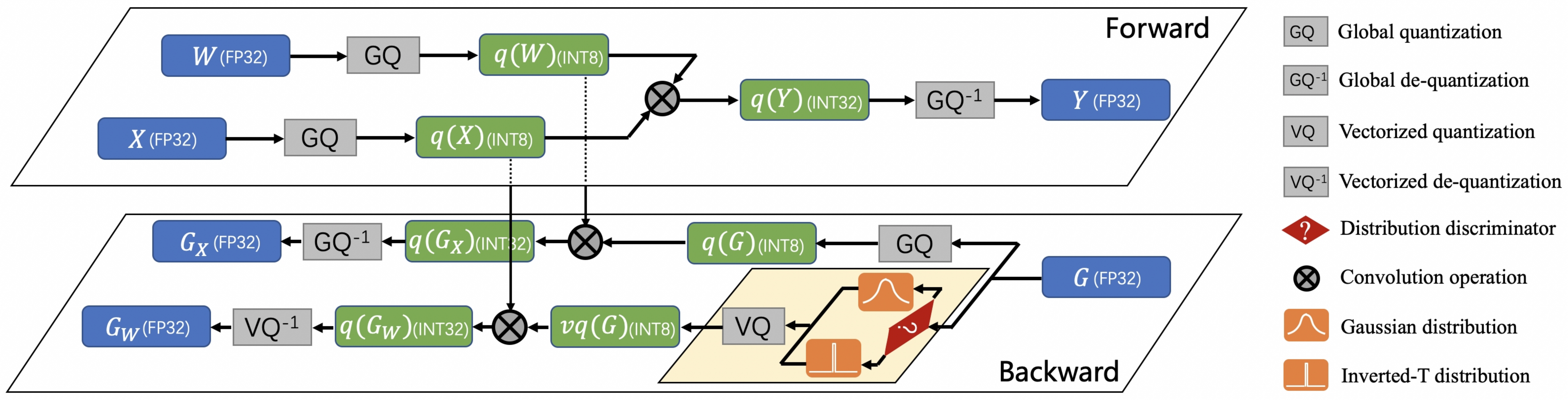} 
	\caption{The dataflow of Distribution Adaptive INT8 Quantization for convolution layer. In the forward pass, we use global quantization to quantize weight and activation. While in the backward process, we adopt vectorized quantization of gradient to calculate $\mathbf{G}_W$, and global quantization for $\mathbf{G}_X$.}
	\label{Fig.main1} 
\end{figure*}

\subsubsection{Inverted-T distribution.} As shown in the Figure \ref{fig.channelperspective}, the majority of gradients are concentrated near zero, while those far away from the center have much smaller density. This extremely unbalanced distribution can be well portrayed by a piecewise uniform distribution, namely:
\begin{align}
p_{T^{-1}}(g) &=
\left\{\begin{array}{l}
a, \ \ \, \, 
\left | g \right |\in (0,\epsilon)\\ 
b,  \quad
\left | g \right | \in (\epsilon,\left| g \right| _{max})
\end{array}\right. \label{eqn.invT},
\end{align}
where $a$ and $b$ determines distribution intensity with $a>>b$, and $\epsilon$ is a small value. Note that $\left| p(g)-p_{T^{-1}}(g) \right| $ is trivial comparing to the large value of $a$.

Combining Eq.(\ref{eqn.invT}) and Eq.(\ref{eqn.e2}), the first order derivative of $E$ with respect to $s$ is reduced to:
\begin{align}
\nonumber
\frac{\partial E}{\partial s} = 
\frac{\left ( a-b \right )e^{\alpha \epsilon }+ 
	b(255+\alpha s)e^{\alpha s} - 
	254be^{\alpha \left| g\right| {max}} - a }{127\alpha}. 
\end{align}
To eliminate the influence of $\alpha$, we make Eq.(\ref{eqn.deri1}) of adjacent iterations equal:
$\left. \frac{\partial E}{\partial s} \right |_{s=\bar{s}_{t}} =
\left. \frac{\partial E}{\partial s} \right |_{s=\bar{s}_{t-1}} = 0 \label{eqn.timef}$
. Then we can derive the following iterative formula:
\begin{align}
\bar{s}_{t} = (1-kA)\bar{s}_{t-1}+A\left | g \right |_{max,t} \label{eqn.fats},
\end{align}
where $k$ and $A$ are the hyper-parameters (The proof of Eq.(\ref{eqn.inequal}) and Eq.(\ref{eqn.fats}) are presented in Appendix). Eq.(\ref{eqn.fats}) indicates that the gradients calculated from a mini-batch can be improved by the complete dataset after sufficient iterations. Therefore, the update formula in Eq.(\ref{eqn.fats}) may have sort of regularization effect on the gradient.

\subsection{Distribution Adaptive INT8 Quantization}
Finally, we show how to distinguish different distributions along channel dimension. Based on our observation, Inverted-T distribution ($T^{-1}$) has more values concentrated around zero than Gaussian distribution. Thus, we propose the following gradient discriminator:
\begin{align}
\left\{\begin{array}{l}
p\sim N(\mu ,\sigma ^2),  P(\left | g \right | > \sigma) > \lambda \\
p\sim T^{-1},    \qquad         P(\left | g \right | > \sigma) \leq \lambda
\end{array}\right. \label{eqn.judge},
\end{align}
where $\mu$ is the mean of gradient. The threshold $\lambda$ is set as 0.3 in this work, because more than 30\% of values drawn from a Gaussian distribution are outside $\sigma$ \cite{patel1996handbook} theoretically. 

We summarize the complete dataflow of applying our INT8 quantization for convolution layer in Figure \ref{Fig.main1}. Two quantizations of gradient (vectorized + global) can be easily optimized with kernel fusion in implementation. To highlight the core pipeline of quantizing gradient, we describe the backward pass of Distribution Adaptive INT8 Quantization in Algorithm \ref{alg:DAIQ}.

\begin{algorithm}[t]
	\caption{Backward pass of Distribution Adaptive INT8 Quantization.}
	\label{alg:DAIQ}
	\begin{algorithmic} [1]
		\Require
		Full precision gradient of $l$-layer $G_{Y_l}$ in $t$ training iteration with shape $(N,C,H,W)$, quantized activation and weight saved in forward pass $q(X_{l-1})$ and $q(W_l)$, and their quantization parameters $s_{X_{l-1}}$ and $s_{W_l}$.
		\Ensure
		$G_{X_{l-1}}$ and $G_{W_l}$ // gradients for $X_{l-1}$ and $W_l$
		\For{$c \leftarrow 1 $ to $C$}
		\State Infer $\left | g \right |_{max,l,c}$ and $\sigma_{l,c}$; 
		\State // density of gradients larger than $\sigma_{l,c}$
		\State $P=\sum_{\left | g \right |> \sigma_{l,c}}^{}\frac{1}{N\cdot H\cdot W}$; 
		\If {$P>\lambda$}  // update as Gaussian distribution
		\State $s_{l,c,t}=\left | g \right |_{max,l,c}$;   
		\Else \ // update as Inverted-T distribution
		\State $s_{l,c,t}=(1-kA)s_{l,c,t-1}+A\left | g \right |_{max,l,c}$; 
		\EndIf
		\EndFor
		\State $vq(G_{Y_{l}})=VQ(G_{Y_{l}}, s_{l})$; // vectorized quantization
		\State $q(G_{Y_{l}})=Q(G_{Y_{l}}, \left | g \right |_{max,l})$; // global quantization
		\State $q\left ( G_{W_{l}} \right )= q\left ( X_{l-1}^{\prime} \right ) \otimes vq\left ( G_{Y_l}^{\prime} \right )$ ; // dilation convolution
		\State $q\left ( G_{X_{l-1}} \right ) = q\left ( G_{Y_l} \right ) \odot q\left ( W_{l} \right ) $; // transpose convolution
		\State // apply vectorized/global de-quantization
		\State $G_{W_{l}} = VQ^{-1}\left ( q\left ( G_{W_{l}} \right ) , s_{W_l} , s_{l} \right )$; 
		\State $G_{X_{l-1}} = Q^{-1}\left ( q\left ( G_{X_{l-1}} \right ) , s_{X_{l-1}} , \left | g \right |_{max,l} \right )$.
	\end{algorithmic}
\end{algorithm}

\section{Experiments}
In this section, we evaluate the proposed method on different tasks and test the training speedup in practice. We use cosine scheduler \cite{loshchilov2016sgdr} for MobileNetV2 \cite{sandler2018mobilenetv2}, and multistep learning rate scheduler for the others. All experimental settings for INT8 training are consistent with the full precision model (baseline). Due to the mismatched baselines between different works, we adopt the accuracy drop ( $\Delta=$ INT8 - baseline) as a metric if not specified.

\subsection{Ablation study}
The ablation study is conducted on ImageNet \cite{deng2009imagenet} with ResNet-18 \cite{he2016deep}. We set the global quantization (GQ) of gradient as baseline, which takes $\left | g \right |_{max}$ as quantization parameter.

\begin{figure} [t]
	\centering
	\includegraphics[width=0.47\textwidth]{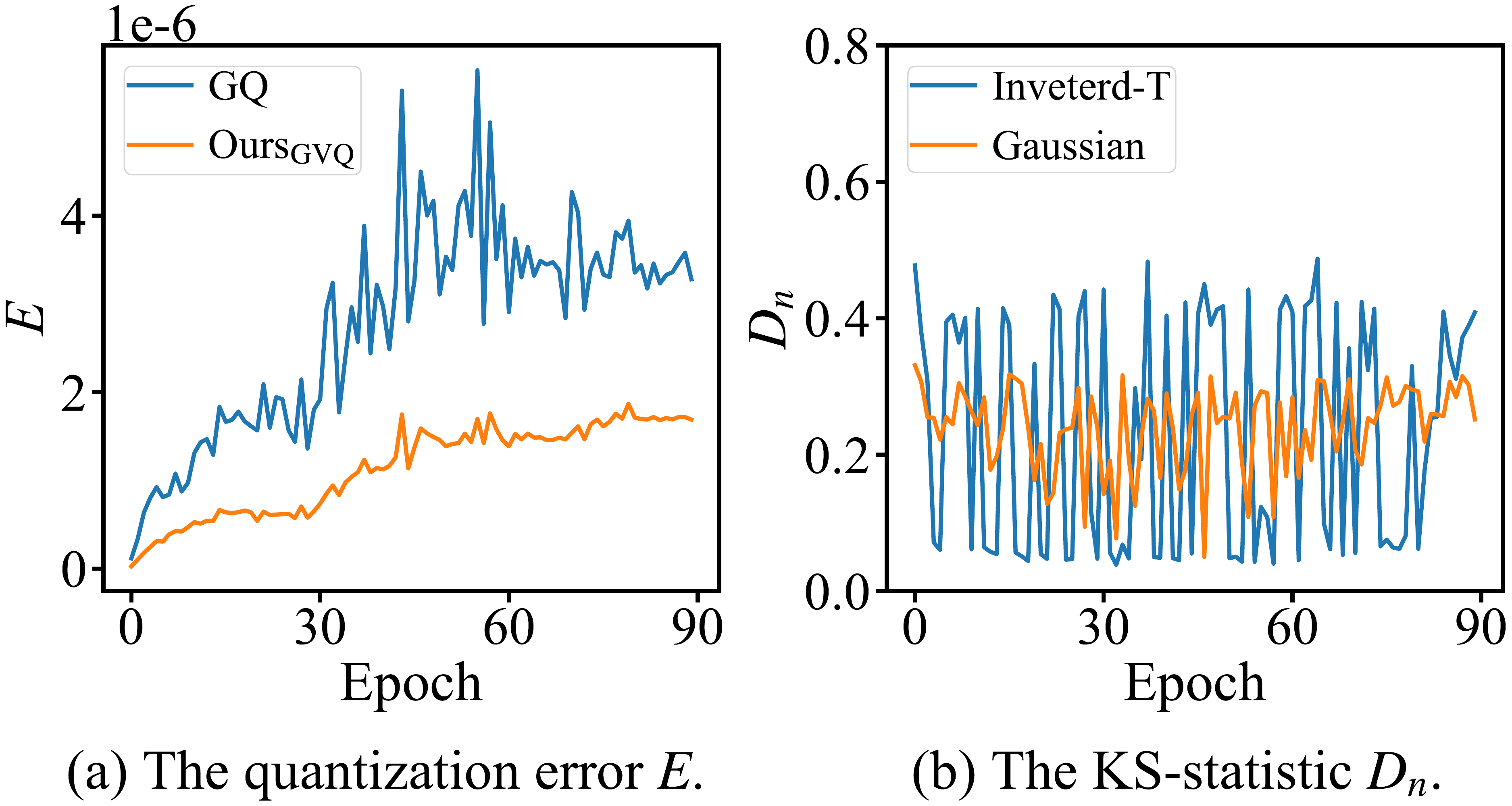} 
	\caption{The evolution curves of quantization error (with $\alpha = 0.2$ for representative) and KS-statistic for a convolution layer in ResNet-18.} 
	\label{Fig.prop} 
\end{figure}

\subsubsection{Gradient Vectorized Quantization (GVQ).} To verify the effectiveness of vectorized quantization, we use the identical strategy of GQ to update the quantization parameter along channel dimension (denoted as Ours$\mathrm{_{GVQ}}$). Figure \ref{Fig.prop} (a) presents the evolution of quantization errors $E$ in different epochs. 
We can find the quantization error of Ours$\mathrm{_{GVQ}}$ is significantly lower than that of GQ during the total training process, which shows the quantization strategy of treating each channel differently is more reasonable than GQ.

\subsubsection{Magnitude-aware Clipping Strategy (MCS).} In the second experiment, we apply MCS to the layer-wise (not channel-wise) gradients based on Eq.(\ref{eqn.judge}) to get free of GVQ (denoted as Ours$\mathrm{_{MCS}}$). We utilize Kolmogorov-Smirnov (KS) statistic \cite{kolmogorov1933sulla} to quantify the difference between two distributions:
\begin{align}
D_n = sup_{g}\left | F_1\left( g \right)  - F_2\left( g \right) \right |,
\end{align}
where $sup_g$ is the supremum of the set of distances, $F_1$ is the cumulative distribution function of gradient, and $F_2$ is for the approximation one (Gaussian/Inverted-T). Figure \ref{Fig.prop} (b) shows the $D_n$ of Inverted-T (blue curve) is lower than that of Gaussian (orange curve) in some intervals of training epochs, indicating the rationality of Inverted-T assumption. Compared with previous methods that assume Gaussian distribution of gradient, our distribution adaptive strategy can describe the ever-changing gradient more precisely. 

\begin{table}[h] 
	\centering
	\begin{tabular}{p{0.9cm}<{\centering} p{0.7cm}<{\centering} p{0.7cm}<{\centering} p{0.7cm}<{\centering} p{0.7cm}<{\centering} p{0.7cm}<{\centering} p{0.7cm}<{\centering} } 
		\toprule 
		$k$ & 1.0  & 1.0 & 1.2 & 1.2 & 1.5 & 1.5 \\  
		$A$ & 0.5  & 0.8 & 0.5 & 0.8 & 0.5 & 0.8\\ 
		\midrule 
		Acc./\%    &   69.94   &  70.01  &  69.90 & 69.82  & 69.80 & 69.91  \\ 
		\bottomrule 
	\end{tabular} 
	\caption{\label{tbl.akinfu} Comparison of different $k$ and $A$.} 
\end{table}

Furthermore, we explore the sensitivity of two additional hyper-parameters ($k$ and $A$) introduced in MCS. As Table \ref{tbl.akinfu} lists, the accuracy between different hyper-parameters are similar, which proves the stability of our MCS. In the following experiments, we set $k=1$ and $A=0.8$.

\subsubsection{Results.} 
Table \ref{tbl.abalation} summarizes the final accuracy of different quantization schemes. Ours$\mathrm{_{GVQ}}$ and Ours$\mathrm{_{MCS}}$ obtain 0.30\% and 0.33\% accuracy improvement over GQ respectively, showing again these two strategies are effective. Not surprisingly, our proposed framework has the highest accuracy improvement, i.e., 0.53\%, which exceeds both Ours$\mathrm{_{GVQ}}$ and Ours$\mathrm{_{MCS}}$. This is attributed to the fact that our classification of gradient distribution from channel perspective has a smaller $D_n$ value than that of layer perspective.

\begin{table}[h] 
  \centering
  \begin{tabular}{ccccc} 
    \toprule 
    Method & GQ & Ours{$\mathrm{_{GVQ}}$} & Ours$\mathrm{_{MCS}}$ & Ours \\
    \midrule 
    Acc / \%& 69.68 & 69.98 & 70.01 & \textbf{70.21} \\ 
    \bottomrule 
  \end{tabular} 
  \caption{\label{tbl.abalation}Ablation study on different quantization strategies.} 
\end{table}

\subsection{Image Classification}
Next, we present the classification performance on CIFAR-10 \cite{krizhevsky2009learning} and ImageNet datasets, with different backbones: AlexNet \cite{krizhevsky2012imagenet}, ResNet, VGG \cite{simonyan2014very}, MobileNetV2, and InceptionV3 \cite{szegedy2016rethinking}. We compare our method with the following state-of-the-art: Unified Int8 training (UI8) \cite{zhu2020towards}, Adaptive Fix-Point training (AFP) \cite{zhang2020fixed}, WAGEUBN \cite{yang2020training}, FP8 training \cite{wang2018training} and DoReFa-Net \cite{zhou2016dorefa}.

\begin{table}[t]
	\centering
	\begin{tabular}{ccccc} 
		\toprule 
		Model & Method & FP32/\%  & INT8/\% & $\Delta$/\% \\ 
		\midrule
		\multirow{2}*{ResNet-20} &
		UI8          & 92.32  & 91.95  & -0.37 \\
		& Ours         & 92.35  & 92.76  & \textbf{0.41} \\
		\midrule
		\multirow{2}*{MobileNetV2} &
		UI8          & 94.39  & 93.38  & -1.01 \\
		& Ours         & 94.73  & 94.37  & \textbf{-0.36} \\
		\midrule
		\multirow{2}*{InceptionV3} &
		UI8          & 94.89  & 95.00  & 0.11 \\
		& Ours         & 95.00  & 95.21  & \textbf{0.21} \\
		\bottomrule 
	\end{tabular} 
	\caption{\label{tbl.cifar2}The Top-1 Accuracy on CIFAR-10 dataset.} 
\end{table}

\begin{table}[t]
	\begin{threeparttable}
		\centering
		
		\begin{tabular}{ p{1.8cm}<{\centering} |p{1.8cm}<{\centering} p{0.9cm}<{\centering} p{0.9cm}<{\centering} p{0.9cm}<{\centering}} 
			\toprule [1pt]
			Model & Method & FP32/\%  & INT8/\% & $\Delta$/\% \\ 
			\midrule [1pt]
			\multirow{6}*{Alexnet} & UI8    &  59.84 &  58.84 & -1.00 \\
			& AFP    &  58.00 &  58.22 & \textbf{0.22} \\
			& DoReFa-Net&  55.90 &  53.00 & -2.90 \\
			& FP8 training&  58.00 &  57.50 & -0.50 \\
			& Ours   &  58.18 &  58.21 & 0.03 \\
			\midrule
			\multirow{3}*{ResNet-18} & UI8    &  70.30 &  69.67 & -0.63 \\
			& WAGEUBN & 68.70     &  67.40 & -1.30 \\
			& FP8 training&  67.43 &  67.34 & -0.09 \\
			& Ours   &  70.22 &  70.21 & \textbf{-0.01} \\
			\midrule
			\multirow{3}*{ResNet-34} & UI8    &  73.68 &  73.29 & -0.39 \\
			& WAGEUBN & 71.99      &  68.50 & -3.49 \\
			& Ours   &  73.46 &  73.40 & \textbf{-0.06} \\
			\midrule
			\multirow{4}*{ResNet-50} & UI8    &  76.60 &  76.34 & -0.26 \\
			& AFP    &  76.40 &  76.20 & -0.20 \\
			& WAGEUBN & 74.66      &  69.07 & -5.59 \\
			& Ours   &  76.50 &  76.59 & \textbf{0.09} \\
			\midrule
			\multirow{3}*{InceptionV3} & UI8    &  72.39 &  71.20 & -1.19 \\
			& AFP    &  73.00 &  72.80 & -0.20 \\
			& Ours   &  75.47 &  75.48 & \textbf{0.01} \\
			\midrule
			\multirow{2}*{VGG-16} & AFP    &  71.00 &  70.60 & -0.40 \\
			& Ours   &  72.39 &  72.44 & \textbf{0.05} \\
			\midrule
			\multirow{3}*{MobileNetV2} & UI8    &  72.39 &  71.20 & -1.19 \\
			& AFP    &  71.80 &  70.50 & -1.30 \\
			& Ours   &  72.44 &  71.92 & \textbf{-0.52} \\
			\bottomrule [1pt]
		\end{tabular} 
	\end{threeparttable}
    \caption{\label{tbl.imagenet}The Top-1 Accuracy on ImageNet dataset.} 
\end{table}

\subsubsection{CIFAR-10.} It’s clear the proposed method has less accuracy drop than UI8 for every examined model, as shown in Table \ref{tbl.cifar2}. For MobileNetV2, we even make the accuracy loss less than 0.4\%, proving the effectiveness of our INT8 training framework. Our method achieves higher training accuracy than baseline for InceptionV3 just like UI8, but we extend it to ResNet-20 further.

\subsubsection{ImageNet.} Table \ref{tbl.imagenet} shows the top-1 accuracy of different backbones on ImageNet dataset. The performance of our INT8 quantization is superior to other state-of-the-art methods on all models. The quantization strategy of WAGEUBN is too aggressive  (quantizing all data paths), resulting in considerable accuracy loss on many models (larger than 5\% on ResNet-50). For AlexNet, DoReFa-Net has the worst performance due to its inappropriate quantization of gradient. FP8 training performs better than other methods on ResNet-18, but it depends on new hardware to boost the speed. For the popular ResNet-50, our method achieves even 0.09\% accuracy improvement over baseline, while UI8 and AFP cause more than 0.2\% accuracy drop, which verifies our multiple quantization parameters are more reasonable than only one. Although AFP has better performance in AlexNet than ours, their computation is more intensive owing to 77.5\% INT16 calculation. 

Because of the sparsely linked depth-wise convolution structure, the quantization of MobileNetV2 in previous works always cause a significant accuracy drop, i.e., 1.19\% for UI8 and 1.30\% for AFP. Equipped with our method, the accuracy drop immediately decreases to 0.52\%, which is almost twice better than UI8 and AFP. Therefore, our consideration of the inherent gradient distributions and magnitudes can benefit the quantization of complex convolution layer.

To the best of our knowledge, except for MobileNetV2, we are the first INT8 framework to obtain almost lossless performance compared with full precision, which is hardly achieved by previous studies. It is very important in practical scenarios, because one prefers spending more time training the FP32 model if there are some loss of accuracy in the INT8 quantization scheme. Moreover, our performance on AlexNet, VGG-16 and InceptionV3 are slightly better than the baseline, which may be attributed to the regularization effect originated from Eq.(\ref{eqn.fats}).

\begin{table}[t] 
	\centering
	\begin{tabular}{p{1.2cm}<{\centering} |p{1.0cm}<{\centering} | p{0.9cm}<{\centering} p{0.9cm}<{\centering} p{0.9cm}<{\centering} p{0.9cm}<{\centering} } 
		\toprule [1pt]
		Dataset & Model & Method & FP32/\%  & INT8/\% & $\Delta$/\% \\ 
		\midrule [1pt]
		\multirow{5}*{COCO} & 
        Faster-  &  UI8            & 36.9  & 35.1  & -1.8 \\
& RCNN    & Ours         & 37.5  & 37.4  & \textbf{-0.1} \\	
          \cmidrule{2-6}	
	    &	Retina  & UI8            & 36.2  & 35.0  & -1.2 \\
        &  -Net  & Ours         & 36.4  & 36.3  & \textbf{-0.1} \\
		\midrule 
		\multirow{5}*{\shortstack{PASCAL\\-VOC}} & Faster- & UI8            & 82.0  & 81.9  & -0.1 \\
		 & RCNN & Ours         & 79.5  & 79.4  & \textbf{-0.1} \\
		\cmidrule{2-6}
		 & Retina &
		UI8            & -  & -  & - \\
		& -Net & Ours         & 77.3  & 77.4  & \textbf{0.1} \\
		\bottomrule [1pt]
	\end{tabular} 
	\caption{\label{tbl.detection}Detection Results on different datasets.} 
\end{table}

\subsection{Object Detection}
To further validate the extensibility of our method, we train Faster-RCNN \cite{ren2015faster} and RetinaNet \cite{lin2017focal} on COCO \cite{lin2014microsoft} and PASCAL VOC datasets \cite{everingham2010pascal}. We adopt the open-source MMdetection framework \cite{mmdetection} with the default settings and take ResNet-50 as backbone. Mean Average Precision (mAP) is used as evaluation metric here. 

As listed in Table \ref{tbl.detection}, our method achieves satisfactory results on both datasets, which are less than 0.1\% accuracy drop compared to full precision model. In particular, we surpasses UI8 on the large-scale COCO dataset by a large margin (more than 1\%). Our good performance on both Faster-RCNN (two-stage) and RetinaNet (one-stage) also reveal the versatility of our method for different types of detectors. In contrast, UI8 suffers from dramatic accuracy drop (larger than 1\%) on these two models. Note that we are the first to report almost lossless INT8 results on the object detection task with various models.

\subsection{Video Classification}
Applying quantization to 3D convolution is very meaningful owing to the high computation complexity of 3D convolution layer. The transition of our method from 2D to 3D convolution layer is straight forward: treating depth dimension in the same way as height and width. In addition, we found the optimal hyper-parameters ($k$, $A$) used in 2D layer performs well in 3D case, showing the stability of our quantization scheme again. We conduct experiments with ResNet3D-18/50 for video classification. Two widely used dataset, i.e., UCF-101 \cite{soomro2012ucf101} with official split-1 and Kinetics-400 \cite{kay2017kinetics} released in 2017 are both tested. 

\begin{table}[h] 
  \centering
  \begin{tabular}{p{1.8cm}<{\centering}|p{2.0cm}<{\centering} p{0.9cm}<{\centering} p{0.9cm}<{\centering} p{0.6cm}<{\centering} } 
    \toprule 
    Dataset & Model  & FP32/\%  & Ours/\% & $\Delta$/\% \\ 
    \midrule 
    \multirow{2}*{UCF-101} &
        ResNet3D-18    & 54.9   & 55.1 & \textbf{0.2}\\
    & ResNet3D-50    & 57.6   & 57.5 & \textbf{-0.1}\\
    \midrule
    \multirow{2}*{Kinetics-400} &
      ResNet3D-18    & 51.0   & 51.4 & \textbf{0.4} \\
    & ResNet3D-50    & 58.5   & 58.8 & \textbf{0.3} \\
    \bottomrule 
  \end{tabular} 
  \caption{\label{tbl.video}The Top-1 Accuracy of Video Classification.} 
\end{table}

Our INT8 quantization in all examined models perform comparable with (even better than) baseline in Table \ref{tbl.video}. The success of our method in 3D convolution demonstrates the similarity between depth and height/width components, which also reflects the importance of our analysis of gradient from the channel dimension.

\begin{figure} [t]
  \centering
  \includegraphics[width=0.47\textwidth]{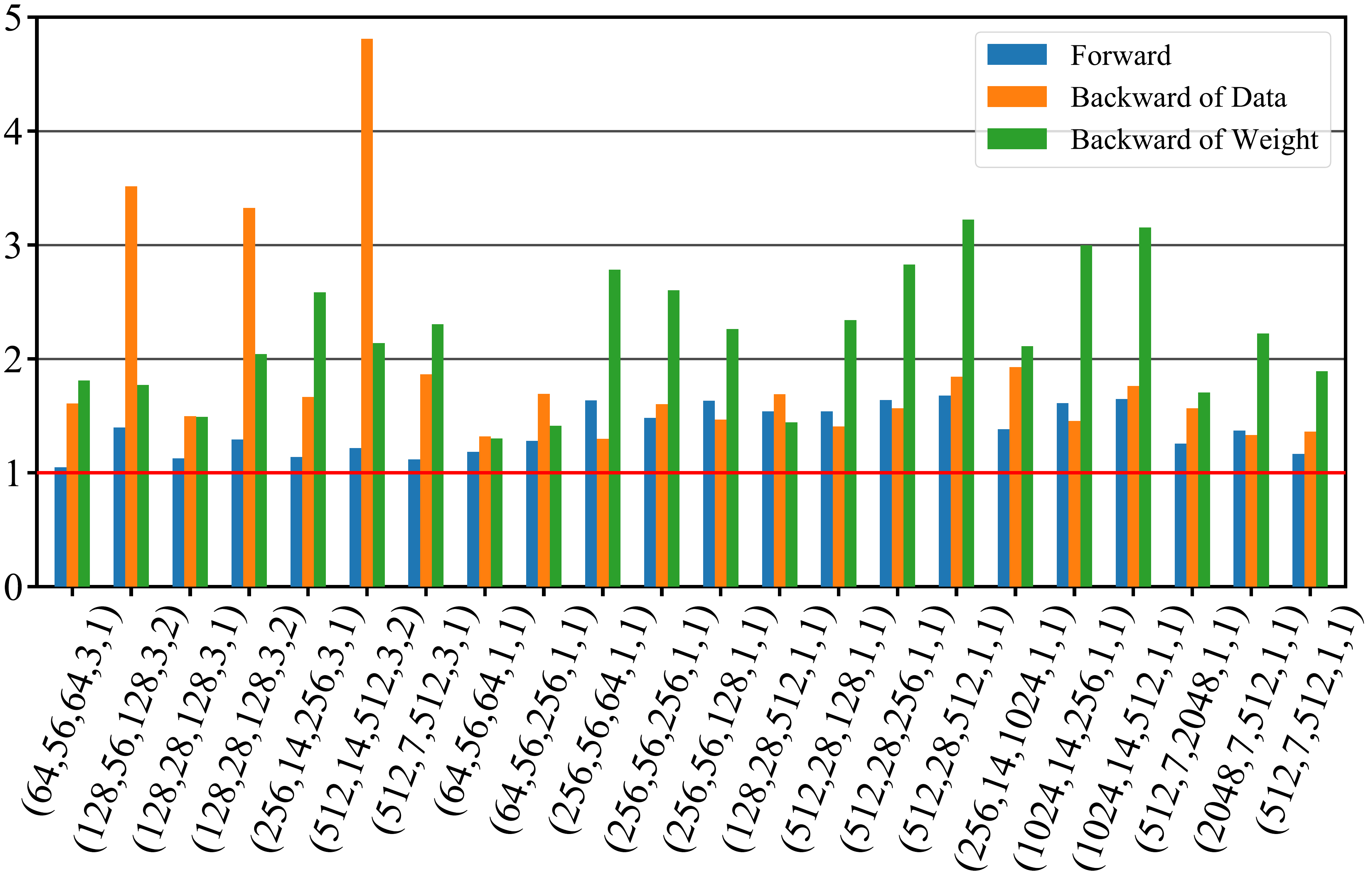} 
  \caption{The speedup of INT8 over FP16 on convolution layer, where X-axis is (channel numbers, feature size, kernel number, kernel size, stride, batch size = 64), and Y-axis is speedup multiple.} 
  \label{Fig.speedup} 
\end{figure}

\subsection{Acceleration Results}
To make full use of the computational capacity of INT8, we take the advantage of TensorCore to implement the convolution kernel. We specially design three-stage pipelines: GPU memory $\Rightarrow$ Shared memory $\Rightarrow$ Register $\Rightarrow$ TensorCore, to make computation and memory access fully parallel (Appendix contains more details).

\begin{table}[t] 
  \centering
  \begin{tabular}{lccc} 
    \toprule 
     Training Process & FP32  & FP16 & INT8  \\  
     \midrule 
     Forward (ms)      &    71.1   & 38.3   & 33.7 \\ 
    Backward (ms)  &  151.3  & 87.4   & 72.6 \\ 
    Iteration (ms)    &  237.8   & 136.4   & 115.6    \\ 
    \bottomrule 
  \end{tabular} 
  \caption{\label{tbl.speedsum} Running time of ResNet-50 on GeForce RTX 2080Ti (batch size = 64).} 
\end{table}

In Figure \ref{Fig.speedup}, we compare our INT8 implementation over FP16 version (based on cuDNN 7.6.5) on various convolution layers of ResNet-50. The implementation algorithm is limited to IMPLICIT\_PRECOMP\_GEMM\footnote{\url{https://docs.nvidia.com/deeplearning/cudnn/api/index.html#cudnnConvolutionFwdAlgo_t}} for fair comparison. Our INT8 kernel is 1.76$\times$ faster than FP16 implementation with kernel size=1, and 1.94$\times$ faster with kernel size=3. Table \ref{tbl.speedsum} shows the average end-to-end time for one iteration of the training. We achieves 18\% faster than FP16 implementation and more than 200\% faster over FP32 version. 

Considering FP16 quantization has been well studied, and supported by NVIDIA, it poses a great challenge to make INT8 quantization even faster in practice. To our best knowledge, we are the first to achieve actual speedup over FP16. We believe that if more engineering efforts are taken with INT8, it will bring more significant performance improvement. 

\section{Conclusion}
In this work, we well capture the gradient distribution as the key to achieve almost lossless quantization training for convolutional neural networks. In particular, we propose \emph{Gradient Vectorized Quantization} to address multiple distributions along channel dimension, and \emph{Magnitude-aware Clipping Strategy} to automatically update the optimal quantization parameters after introducing effect of gradient magnitude. Our approach has successfully trained on a lot of networks and deep learning tasks with negligible accuracy degradation, which sets a new state-of-the-art to the community. With the kernel level optimization, we further achieve more than 200\% acceleration on modern Turing architecture.

\bibliography{ref}
\end{document}